\def\year{2020}\relax
\documentclass[letterpaper]{article} 
\usepackage{aaai20}  
\usepackage{times}  
\usepackage{helvet} 
\usepackage{courier}  
\usepackage[hyphens]{url}  
\usepackage{graphicx} 
\usepackage{graphicx}  
\usepackage{mathtools}
\usepackage{amsfonts,amssymb}
\usepackage{amsmath}
\newtheorem{myDef}{Definition}
\newtheorem{myExa}{Example}
\usepackage{graphicx}
\usepackage{subcaption}
\usepackage[noend]{algpseudocode}
\usepackage{algorithmicx,algorithm}
\usepackage{multirow}
\usepackage{booktabs}
\title{Estimating Early Fundraising Performance of Innovations\\ via Graph-based Market Environment Model }
\author{
Likang Wu\textsuperscript{\rm 1},
Zhi Li\textsuperscript{\rm 1},
Hongke Zhao\textsuperscript{\rm 2},
Zhen Pan\textsuperscript{\rm 1},
Qi Liu\textsuperscript{\rm 1},
Enhong Chen\textsuperscript{\rm 1}\thanks{Corresponding Author.}\\
\textsuperscript{\rm 1}Anhui Province Key Laboratory of Big Data Analysis and Application, University of Science and Technology of China\\
\textsuperscript{\rm 2}College of Management and Economics, Tianjin University\\
\{wulk, zhili03, pzhen\}@mail.ustc.edu.cn, \{qiliuql, cheneh\}@ustc.edu.cn, hongke@tju.edu.cn
}
 
\begin{document}

\maketitle

\begin{abstract}
\textit{Well begun is half done}. In the crowdfunding market, the early fundraising performance of the project is a concerned issue for both creators and platforms. However, estimating the early fundraising performance before the project published is very challenging and still under-explored. To that end, in this paper, we present a focused study on this important problem in a market modeling view. Specifically, we propose a Graph-based Market Environment model (GME) for estimating the early fundraising performance of the target project by exploiting the market environment. In addition, we discriminatively model the \textit{market competition} and \textit{market evolution} by designing two graph-based neural network architectures and incorporating them into the joint optimization stage. Finally, we conduct extensive experiments on the real-world crowdfunding data collected from Indiegogo.com. The experimental results clearly demonstrate the effectiveness of our proposed model for modeling and estimating the early fundraising performance of the target project.
\end{abstract}

\section{Introduction}
Recent years have witnessed the rapid development of online crowdfunding platforms, such as Indiegogo~$\footnote{https://www.indiegogo.com/}$ and Kickstarter~$\footnote{https://www.kickstarter.com/}$. More and more entrepreneurs and individuals create projects on these platforms to demonstrate their innovative ideas and solicit funding from the public.

The prevalence of crowdfunding platforms triggers many research problems, such as project success prediction~\cite{li2016project}, recommendations~\cite{zhang2019personalized} and funding dynamic tracking~\cite{zhao2017tracking,zhao2019voice}. Most of these existing studies focus on modeling the fundraising process after the project published. However, for creators and platform operators, they all want to understand and estimate the early fundraising performance before the project setup. Actually, according to the official statistics of online crowdfunding platforms, the fundraising ability of the project reached its peak in the first 24 hours~$\footnote{https://pse.is/KWP5T}$, and projects with early funding achieve 30\% of corresponding goals are most likely to succeed~$\footnote{https://www.startups.com/library/expert-advice/key-crowdfunding-statistics}$. But it is hard for the founder to ensure that the project has a good start due to the complex and dynamic market environment. Thus, modeling the market environment is an essential issue for judging whether it is a good time to start the project. Unfortunately, less effort has been made towards this goal. To that end, in this paper, we conduct a focused study on modeling and estimating the early performance of innovations in a market view. To the best of our knowledge, this is the first attempt in this area.

Indeed, it is very challenging to estimate the early performance of unpublished innovations. As mentioned above, the project fundraising performance is largely influenced by the market environment. On the one hand, once a project is published, it will face market competition from other projects after published. That is difficult to model the complex competition relationships and estimate the competitiveness of the target project. On the other hand, market evolution can also have a huge impact on the fundraising process. Then, tracking market evolution is another challenge for estimating the early performance of the target innovation.

To address the challenges mentioned above, in this paper, we present a focused study on exploiting the project early fundraising performance in a market modeling view. Along this line, we propose a novel model, i.e., Graph-based Market Environment model (GME), to explore the market environment of online crowdfunding platforms and estimate the early performance of the target project. Specifically, GME consists of two components, i.e., Project Competitiveness Modeling (PCM) and Market Evolution Tracking (MET). In the PCM module, we apply a hybrid Graph Neural Network (GNN) to model the competitiveness pressure of the target project by aggregating competitive influences of other competitive projects in the market. Additionally, in the MET module, we design a propagation structure to track the market evolution and apply a Gated Graph Neural Networks (GG-NNs)~\cite{li2015gated} to model the evolution process. Next, we optimize our GME with a joint loss to learn the fundraising performance of competitive projects and estimate the early performance of the target project, simultaneously. Finally, we conduct extensive experiments on the real-world crowdfunding data collected from Indiegogo.com. The experimental results clearly demonstrate the effectiveness of our proposed model for modeling and estimating the early fundraising performance of the target project.

\section{Related Work}
In this section, we review the research from two categories, i.e., crowdfunding and graph neural networks.

\subsection{Crowdfunding}
The research of online crowdfunding can be divided into two categories according to the research perspective: for the individual projects and for the entire market. For the individual projects, most researchers paid much attention to the prediction of project success~\cite{li2016project,jin2019estimating}. Besides,~\citeauthor{liu2017enhancing}~\shortcite{liu2017enhancing} optimized the algorithm of production supply to reduce the redundant losses of creators, and ~\citeauthor{zhao2017tracking}~\shortcite{zhao2017tracking} focused on tracking the dynamics for projects in their complete funding durations. For the entire market, ~\citeauthor{lin2018modeling}~\shortcite{lin2018modeling} modeled dynamic competition on crowdfunding markets,~\citeauthor{zhang2019personalized}~\shortcite{zhang2019personalized} are devoted to the recommender systems on the crowdfunding market, and~\citeauthor{janku2018successful}~\shortcite{janku2018successful} propose that good early performance on the crowdfunding market is especially significant.

In spite of the importance of previous studies, they mainly focused on modeling the market reaction or investors' interest after projects launched, and were still lack in a deep and quantitative exploration of the early performance before the project setup. To the best of our knowledge, in this paper, we are the first comprehensive attempt to exploit the prediction of early performance in a market modeling view.

\subsection{Graph Neural Network}
Graph Neural Network (GNN) has been proven successful in modeling the structured graph data due to its theoretical elegance~\cite{bronstein2017geometric}. Extending neural networks to work with graph-structured data was first proposed by ~\citeauthor{gori2005new}~\shortcite{gori2005new}. With the proposal of spectral GNN~\cite{bruna2013spectral} and its improvements ~\cite{defferrard2016convolutional,kipf2016semi}, the research in this area began to develop rapidly. Besides,~\citeauthor{zhou2007learning}~\shortcite{zhou2007learning} regularized the learning procedure of the target task with a hypergraph that captures the higher-order relations among entities.~\citeauthor{gilmer2017neural}~\shortcite{gilmer2017neural} proved that applications of the neural networks to graphs could be seen as specific instances of a learnable message passing framework on graphs.

These years, graph neural network has been applied to many tasks without explicit graph structure. For instance, ~\citeauthor{cucurull2019context}~\shortcite{cucurull2019context} addressed the compatibility prediction problem using a graph neural network that learns to generate product embeddings conditioned on context. ~\citeauthor{feng2019temporal}~\shortcite{feng2019temporal} created the influence graph model by building the edges between related stocks in the stock prediction task. In this paper, we organize the projects in the crowdfunding market as a market graph. And we further propose the GNN-based networks to model the market competitiveness and market evolution, which is the first attempt in this domain.

\section{Methodology}
In this section, we propose a framework to estimate the project's early fundraising performance in a market modeling view. First, we introduce the research problem and the constructed features we use. Then, we present the technical details of our proposed model, i.e., Graph-based Market Environment model (GME).

\subsection{Preliminaries}
In online crowdfunding services, different start-up time has a great impact on the early fundraising performance due to the different market environments~\cite{janku2018successful}. Along this line, we aim at estimating the early performance in a market modeling view. However, the early fundraising amounts are improper to directly measure the early performance, since the same fundraising amount in early-stage means different performance for projects with different fund goals. So in this paper, we use the achievement of goal in the early stage to measure our target value. More specifically, For the target project $i$, we can get the quantitative definition of the early performance $y_i$ as:

\begin{figure*}[t]
\centering
\includegraphics[width=.95\textwidth]{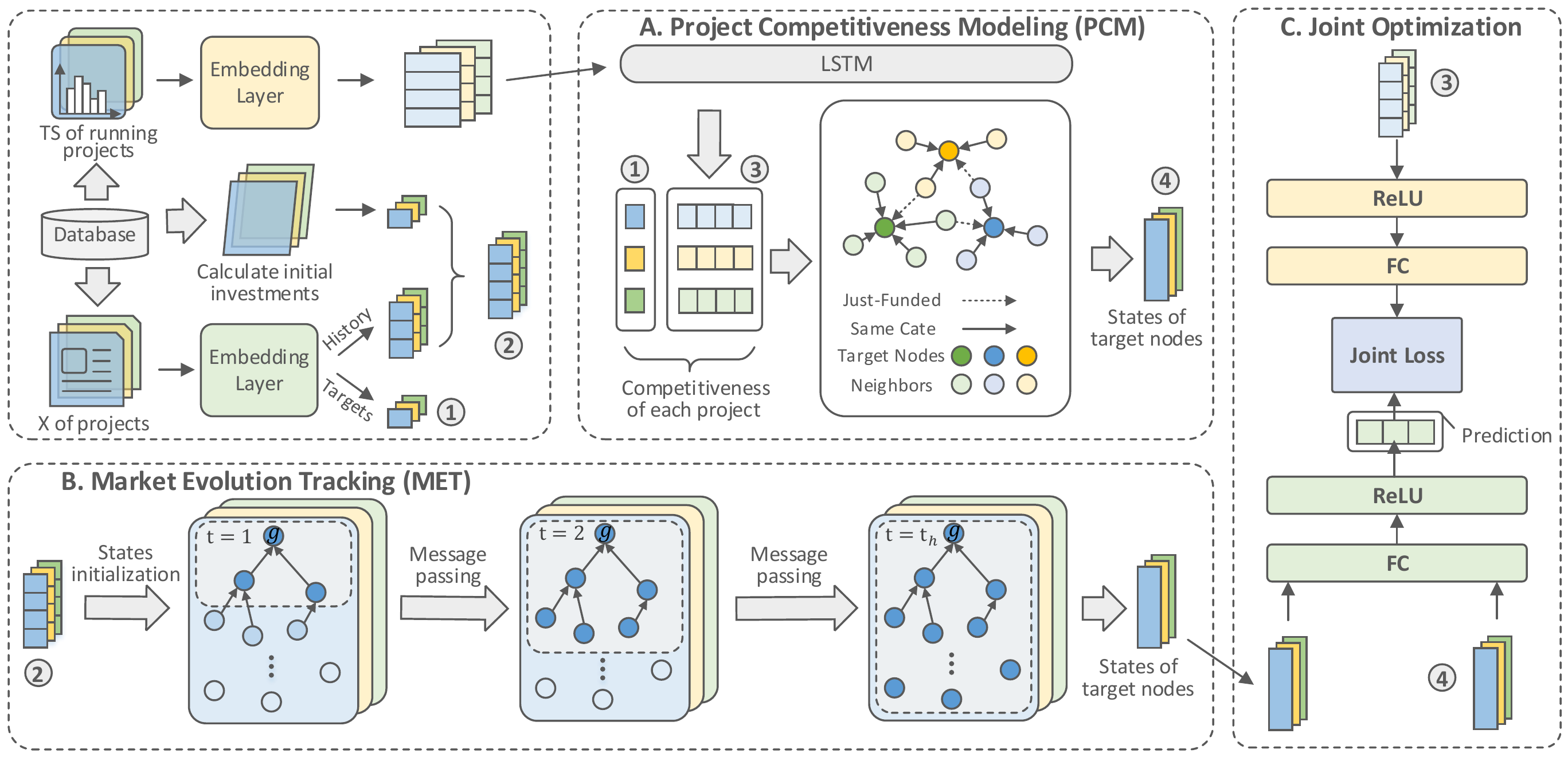} 
\caption{The overview of Graph Market Environment model (GME) : A. Project Competitiveness Modeling (PCM) captures the competitive relationships
between the target projects and others. B. Market Evolution Tracking (MET) tracks the market evolution. C. Joint Optimization jointly trains the model with the joint loss of target predicted loss and competitiveness loss.}
\label{fig:framework}
\end{figure*}

\begin{align}
\label{eqy} y_i = log_2\left(1 + \frac{\alpha_i}{g_i}\right),
\end{align}
where $y_i$ denotes the normalized value, $\alpha_i$ is the amount of soliciting funding during the first 24 hours, $g_i$ is the declared pledged goal, so $\frac{\alpha_i}{g_i}$ represents the percentage of funding in early stage to declared pledged goal.
There are two reasons for using function $log_2(\cdot)$ to normalize $y_i$: one is suppress the large variations between minimum and maximum~\cite{wang2018learning}, the other is that the original percentage values mostly distribute in a small value range, such as $\left[0,\ 0.5\right]$, and the $log_2(\cdot)$ is an increasing function with a decreasing slope which can map small value intervals to larger value intervals to make more discriminative prediction.

Next, the studied problem can be defined as follows:	

\begin{myDef}
	\textbf{Early Fundraising Performance Estimating}. Given the target project $g$ which may be launched at time $T_{g}$ and the static features $\mathbf{X}^{T_{g}}$ of all the related projects with the dynamic fundraising sequence $S^{T_{g}}$ in the market before $T_{g}$, our task is to estimate the early performance $y_g$ for target project $g$.
\end{myDef}

\begin{table}[t]
	\caption{The information of static features.}\smallskip
	\centering
	\resizebox{.73\columnwidth}{!}{
		\smallskip\begin{tabular}{c|c}
			\hline
			\textbf{Features} & \textbf{Type}\\
			\hline
			Project Description & Text\\
			Project Category & Categorical\\
			Creator Type & Categorical\\
			Currency & Categorical\\
			Declared Funding Duration & Numerical\\
			Declared Pledged Goal & Numerical\\
			\hline
		\end{tabular}
	}
	\label{table:features}
\end{table}
\subsubsection{Static Features.}
Given $N$ projects in the whole dataset, the static feature $\mathbf{x}_i \in {\mathbb R}^{m}$ of project $i$ concatenates the vector of basic information published by creators such as project descriptions, perks information, pledged goals, etc. Table~\ref{table:features} lists the detailed information of static features. Specifically, we discretize numerical values by one-hot encoding to categorical type. And, following the example of ~\citeauthor{jin2019estimating}~\shortcite{jin2019estimating}, text data is converted into numerical vectors by the method of doc2vec~\cite{le2014distributed}. Moreover, we use an instance matrix $\mathbf{X} = {\left[\ \mathbf{x}_1, \mathbf{x}_2,...,\mathbf{x}_N \ \right]}^{\mathrm{T}}$ to present all the existing projects and each row is one project feature.

\subsubsection{Market Dynamic Features.}
Unlike the traditional prediction like success prediction based on the historical time series of the target project, we cannot do that since our task reduce the risk before the target project publish. But we need to model the market evolution, the historical fundraising processes of the projects in the market are important to reflect the project competitiveness and market evolution. For the fundraising process of project $i$ in the market, we can generate a fundraising sequence $S_i = \{ \langle v_1, t_1 \rangle , \langle v_2, t_2 \rangle , ... , \langle  v_{\vert S_i \vert}, t_{\vert S_i \vert} \rangle \}$, where $v$ presents the investment amount of the project at time $t$.

\subsection{Project Competitiveness Modeling}

Once a project is published, it would face fierce competition in the market. Then, how to model the competitiveness pressure of the target project in the current market environment is particularly essential. Along this line, we propose the Project Competitiveness Modeling (PCM) module to model the current competitive environment. Specifically, we first utilize sequential modeling to predict the funding status as the competitiveness of published projects in the market. Then, we propose a competitiveness graph to aggregate the competitive influence of published projects.

In detail, we first select the competitive projects. Considering the fund resource in the market is limited, all the funding projects are competitive with each other~\cite{lin2018modeling}. Thus, given the target project $g$, whose funded time is $T_g$. For each project $i$ running on the platform at time $T_g$, we quantify the competitiveness of $i$ as:
\begin{align}
\mathbf{h}_{i}^{c} = \mathrm{LSTM}\left(\mathbf{TS}_i\right),\ i \in {\Psi}_g ,
\end{align}
where we use Long Short-Term Memory networks (LSTM)~\cite{hochreiter1997long} to predict the funding status of each project in the future, ${\Psi}_g$ is the set of running projects.
$\mathbf{TS}_i=\left[\ {\xi}_0, {\xi}_1,..., {\xi}_{23}\ \right]$ denotes the time series of project $i$ during the past 24 hours before $T_{g}$. Since the investment sequence of project $i$ is
$S_i = \left\{ \langle v_1, t_1 \rangle , \langle v_2, t_2 \rangle , ... , \langle v_{\vert S_i \vert}, t_{\vert S_i \vert} \rangle \right\}$, so the hourly amount is:
\begin{align}
\begin{split}
{\xi}_k = log_{2}\left(\sum v_j\right),\\
\end{split}
\end{align}
where $v_j \in S_i$, $T_{g}-(k+1)*\Delta \leq t_j < T_{g}-k*\Delta$, $k = 0,1,...,23$, $\Delta$ represents the time interval of one hour.
To build the influence relations from running projects to the target project, we construct all the directed edges $\langle i, g \rangle$, $i \in {\Psi}_g$, and apply the embedding vectors $\mathbf{h}_{i}^{c}$ as the state of $i$. Note that, the state of node $g$ is generated from full-connected layer with $\mathbf{x}_g$, since $g$ does not have fundraising time series.

Considering the computational pressure of the platform, it is not necessary to meet the demands of creators for precise time. Along this line, we segment one day into six periods according to the general lifestyle of humans~\cite{wu2017sequential}, i.e. ``8:00$\sim$12:00'', ``12:00$\sim$14:00'', ``14:00$\sim$17:00'', ``17:00$\sim$20:00'', ``20:00$\sim$24:00'' and ``0:00$\sim$8:00''. We define a $\textit{target set}$ $\mathcal{G}$ which contains unpublished projects whose pre-funded time in the same period of one day, and we train the target projects in $\mathcal{G}$ simultaneously on one graph. Note that, to prevent the leakage of information, that is, the information after $T_g$ is used in advance, the current timepoint of the target set $\mathcal{G}$ is defined as $T_{\mathcal{G}} = min\left\{ T_i \mid i \in \mathcal{G} \right\}$ when calculating the competitiveness.

However, there is a serious problem that when the running projects set $\Psi$ is large, the algorithm is time-consuming for modeling all the time series by LSTM. To solve this problem, we propose a pruning method.
We think that for a project that just starts less than 24 hours, it would compete for one investor's attention with other projects that are also observed by this investor.
And, the investors are most likely to find the newly created projects in the \textit{just-funded} section and the corresponding \textit{category} section. Because on these sections with
specific restrictions, our target projects is most likely to appear at the first few pages.
According to these, we define the adjacent matrix $\mathbf{A}^G\in{\mathbb R}^{\vert {\mathcal{G}} \vert \times {\vert {\Psi}_g \vert}}$ as:
\begin{align}
\mathbf{A}^{G}_{{\mathcal{M}}_i,{\mathcal{M}}_j} = \left\{
\begin{aligned}
&1,\ \textrm{if}\ T_i - T_j \leq 3\ \textrm{days}\\
&1,\ \textrm{if}\ C_i = C_j\\
&0,\ \textrm{otherwise},
\end{aligned}
\right.
\end{align}
where $\mathcal{M}$ is a function which maps the original id of a project to the unique column id in $\mathbf{A}^G$ between 0 and $\vert {\Psi}_g \vert$, $\mathbf{A}^{G}_{{\mathcal{M}}_i,{\mathcal{M}}_j}$ indicates whether $j$ is connected to $i$, $C_i$ and $C_j$ express the categories of $i$ and $j$. Simply, the pruning approach not only reduce the number of time series modelled by LSTM but also suppress some noise, and we would design ablation experiment to verify this approach can promote the performance of our model.

Due to the project with strong fundraising ability or it's content closer to $g$ may have a greater impact on the target project~\cite{lin2018modeling}, we implement the method like Graph Attention Network (GAT)~\cite{velivckovic2017graph} to aggregate neighborhoods information of the target node $g$ as:
\begin{align}
e_{gi} &= \mathbf{V}^\mathrm{T}\left[\ \mathbf{W}\mathbf{x}_g \parallel \mathbf{W}\mathbf{x}_i\ \right]  ,\\
 \begin{split}
 {\alpha}_{gi} & = {\mathrm{softmax}}\left(e_{gi}\right) \\
 & = \frac{{\mathrm{exp}}\left({\mathrm{LeakyReLU}}\left(e_{gi}\right)\right)}{\sum_{j \in \mathcal N_g } {\mathrm{exp}}\left({\mathrm{LeakyReLU}}\left(e_{gj}\right)\right)},
 \end{split}
\end{align}
where $\cdot ^\mathrm{T}$ represents transposition, $\mathcal N_g$ extract from $\mathbf{A}^{G}_{\mathcal{M}_g:}$ (all nodes pointing to node g) are the neighbors of $g$, and $\parallel$ is the concatenation operation. ${\alpha}_{gi}$ obtained as the normalized attention coefficients are used to compute a linear combination of the features corresponding to them, to serve as the final output features for node $g$:
\begin{align}
\mathbf{H}_{g}^{c} = \sum_{i \in \mathcal N_g}\left({\alpha}_{gi}\mathbf{W}_{h}\mathbf{h}_{i}^{c}\right) ,
\end{align}
where ${\alpha}_{gi}$ is calculated from static content features, but unlike GAT, we use the product of attention weights ${\alpha}_{gi}$ and estimated financing state $\mathbf{h}_i^{c}$ not $\mathbf{x}_i$. In this way, we can give consideration to two influences include fundraising ability and contents of projects at the same time.
\subsection{Market Evolution Tracking}
After modeling the projects competitiveness, we then consider the market evolution process. Actually, the market environment is dynamic and the investment attention is changed over time~\cite{zhao2017sequential}. For example, in summer, the investors may pay more attention to the innovation of intelligent fan and in winter, the attention of investors may change to self-heating wear. Along this line, we design a Market Evolution Tracking (MET) module to model the dynamic evolution of the crowdfunding market. More specifically, since the number of published projects in the historical market can reach hundreds in a few days, using chained sequential models like LSTM to model the dynamic environment with such a large step size would lead to the performance degradation. Therefore, we design a propagation tree-structured GNN with message passing to track the evolution of market.

\begin{figure}[t]	
	\centering
	\begin{subfigure}[t]{.34\columnwidth}
		\centering
		\includegraphics[width=\columnwidth]{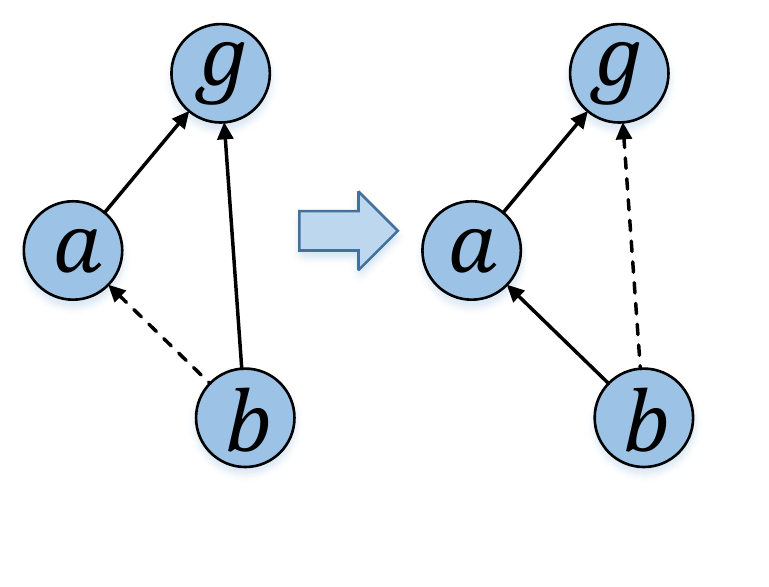}
		\caption{}\label{fig:messa}		
	\end{subfigure}
	\quad
	\begin{subfigure}[t]{.58\columnwidth}
		\centering
		\includegraphics[width=\columnwidth]{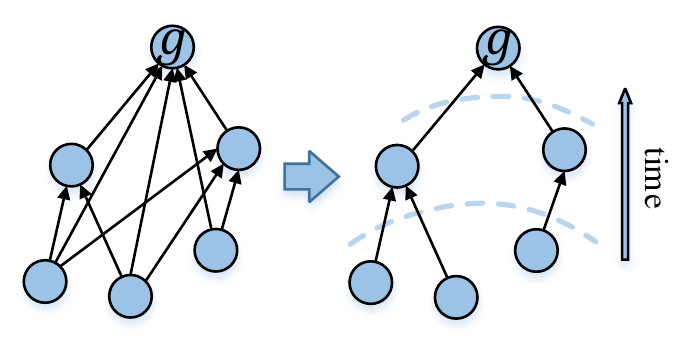}
		\caption{}\label{fig:messb}
	\end{subfigure}
	\caption{The propagation tree structure with message passing. Node $a$ and $b$ are the observable nodes of target $g$.}\label{fig:message}
\end{figure}

Indeed, market environment is the context of a project in the market. To consider the impact of context on fundraising process, for the project $i$ in market, we should refer to the early fundraising states of other projects in historical market. Therefore, the state of published project $j$ is initialized as $\mathbf{h}_{j} = \left[\ \mathbf{x}_{j} \parallel \ {r}_{j}\ \right]$,
$r_j$ is the amount of early fundraising of $j$,
\begin{align}
{r}_j = log_2\left(1 + \sum_{t_l < T_j + n_h\Delta}{v_l}\right),\ v_l \in S_j,
\end{align}
where $n_h$ denotes 24 hours of one day, and there is a constraint: $T_{i} - T_{j} > n_h*\Delta$. Only in this way $i$ can learn the early fundraising performance of published project $j$ by the message propagation through the edge $\langle j, i \rangle$, we call $j$ the \textit{observable node} of $i$.
If the historical data contains past $t_h$ days, the set of published projects is ${\Phi}_i = \{ j \mid j<N,  n_h \ast \Delta  <T_i - T_j< n_h \ast t_h \Delta \}$. Easily the simplest structure to refer to historical market is to calculate the early funding of all nodes in ${\Phi}_i$ firstly, and then create the directed edges from nodes in ${\Phi}_i$ to $i$.
But there is an irrational issue is that we put all the observable projects at the same time level. In an actual environment, there are long-term and short-term effects in forecasting task, the reference values for states with different distances to the pre-startup timepoint should be different. In this paper, we apply the tree-structured graph neural network with message passing to deal with the problem.
\begin{myExa}
To illustrate how to represent the different time levels of projects in the market, let us consider the example in Figure~\ref{fig:messa}. When we considering all the edges between nodes and their observable nodes, there are three edge pairs $\langle a, g \rangle$, $\langle b, g \rangle$, $\langle b, a\rangle$, the length of each edge is greater than $n_h$ hours. If we delete the edge $\langle b, g \rangle$, the depths of $a$, $b$ are 1 and 2 on the $tree_g$ with root $g$, which can represent the different time spans to $T_g$. Moreover, the process of message passing from node $b$ to $a$ to $g$ similar to the step-by-step passing of LSTM.
\end{myExa}

Considering a more complex situation, in Figure~\ref{fig:messb}, we apply the idea in Example 1 to all the observable nodes of target $g$. To prevent information redundancy caused by the states of sub-nodes propagate to $g$ through multiple paths, we build the tree structure with root $g$. This similar idea is appeared in Tree-Lstm to predict the semantic relatedness of sentences in NLP tasks~\cite{tai2015improved}. Our Propagation Tree Construction algorithm is shown in \textbf{Algorithm\ 1}. In this algorithm, we train the target set $\mathcal{G}$ simultaneously as in the competition module, and each $T_g$ is $\textrm{min}\{T_i| i \in \mathcal{G}\}$. So, each $tree_g$ has the same structure except for the root (target node). The adjacency matrix $\mathbf{\Gamma}$ of Algorithm 1 represents the tree structure.
The depths of nodes represent different time levels. Since the market environment is the context of target projects, and the equal interval sampling is beneficial for long-term effects propagation in contextual history learning\cite{wu2017sequential}. Therefore, when creating the edges between nodes in our algorithm, for $i \in U^{k}-U^{k-1}$, we reserve the shortest edge in $\{\langle j, i \rangle \mid \forall j \in U^{k-1} \}$ to build the tree. In this way, the time intervals between the layers of this tree are as similar as possible, it leads to each chain represented by the path from one leaf to root that approaches to evenly spaced time series.

For the message passing of the tree-structured network, we use propagation model in GG-NNs which unroll the recurrence for fixed number of steps $t_h$ and use back propagation through time to compute gradients. We initialize the node state when the time order $t = 1$:
\begin{align}
\begin{split}
\mathbf{h}_{i}^{1} &= \left[\ \mathbf{x}_{i} \parallel r_i\ \right],\ i \in \Phi,\\
\mathbf{h}_{g}^{1} &= \left[\ \mathbf{x}_{g} \parallel 0\ \right],\ g \in \mathcal{G}.
\end{split}
\end{align}
\begin{algorithm}[t]
\caption{Propagation Tree Construction}
{\bf Input:} \\
\hspace*{0.23in}Set of target projects $\mathcal{G}$;\\
\hspace*{0.23in}Set of published observable projects $\Phi$;\\
\hspace*{0.23in}Historical days $t_h$;\\
\hspace*{0.23in}Mapping function ${\mathcal{M}}_i \to \rm{unique\ id}\ in\ \{0,1,...,\vert \Phi \cup \mathcal{G} \vert \},\ i < N $;  \\
{\bf Output:} \\
\hspace*{0.23in}Output adjacency matrix $\mathbf{\Gamma}$
\begin{algorithmic}[1]
\State Set matrix $\mathbf{\Gamma}$ zeros, define infinite matrix $\mathbf{Len}$
\State $U^{1} \gets \mathcal{G}$
\For{$k=1,...,t_h$}
    \State $B \gets \Phi \cup \mathcal{G} - U^k$
    \For{$i \in B$}
        \For{$j \in U^k$}
            \If{$n_h * \Delta  < T_j - T_i < n_h * 2\Delta$}
            	 \State $\mathbf{Len}_{j,i} \gets T_j - T_i$
                \If{$k=1$}
                    \State $\mathbf{\Gamma}_{{\mathcal{M}}_j,{\mathcal{M}}_i} \gets 1$ ~~$\rhd$ connect with roots
                \EndIf
            \EndIf
        \EndFor
        \State $ J \gets {\arg\min}_{j\in U^k} (\mathbf{Len}_{j,i})$
        \State $\mathbf{\Gamma}_{{\mathcal{M}}_j,{\mathcal{M}}_i} \gets 1$
        \State $U^{k+1} \gets U^{k} \cup i$
    \EndFor
\EndFor
\State \Return $\Gamma$
\end{algorithmic}
\end{algorithm}

And in the procedure of message passing, since the target nodes and other nodes update in the same way, we use $v$ denotes the id of each node. The information aggregation in each time step as:
\begin{align}
\begin{split}
\mathbf{a}_{v}^{t} = \mathbf{\Gamma}_{\mathcal{M}_v:}^{\mathrm{T}}\left[\ \mathbf{h}_{1}^{(t-1)\mathrm{T}}...\mathbf{h}_{\vert G \cup \Phi \vert}^{(t-1)\mathrm{T}}\ \right] + \mathbf{b}. \\
\end{split}
\end{align}

We implement gated recurrent units to guarantee the long-step propagation of information, and the gated module for each node at $t$ as follow:
\begin{align}
\begin{split}
\mathbf{z}_{v}^{t} &= \sigma\left(\mathbf{W}_{z}\mathbf{a}_{v}^{t} + \mathbf{U}_{z}\mathbf{h}_{v}^{t-1}\right), \\
\mathbf{r}_{v}^{t} &= \sigma\left(\mathbf{W}_{r}\mathbf{a}_{v}^{t} + \mathbf{U}_{r}\mathbf{h}_{v}^{t-1}\right), \\
\mathbf{h}_{v}^{t'} &= {\mathrm{tanh}}\left(\mathbf{W}\mathbf{a}_{v}^{t} + \mathbf{U}\left(\mathbf{r}_{v}^{t} \odot \mathbf{h}_{v}^{t-1}\right)\right),\\
\mathbf{h}_{v}^{t} &= \left(1 - \mathbf{z}_{v}^{t}\right) \odot \mathbf{h}_{v}^{t-1} + \mathbf{z}_{v}^{t} \odot \mathbf{h}_{v}^{t'}.
\end{split}
\end{align}

After $t=t_h$, we get the final state of target $g$ which learns the influence of propagating from entire market,
\begin{align}
\begin{split}
\mathbf{H}_{g}^{e} &= \mathbf{h}_{g}^{t_h},\ g \in \mathcal{G}.\\
\end{split}
\end{align}

To sum it up, comparing with the chain-like sequential models restricted by time step size, the layers of tree structure is much less, it can model the influence propagation of entire market. Therefore, our propagation tree structure is more suitable for modeling the evolution of entire market.

\subsection{Joint Optimization}
We combine the outputs of the above two modules and design the corresponding loss function. First, we use the fully connected layer to get the output value:
\begin{align}
\begin{split}
\widetilde{y_{g}} = {\mathrm{ReLU}}\left(\mathbf{W}_{f}\left(\mathbf{H}_{g}^{c} + \mathbf{H}_{g}^{e}\right) + b_f\right),\ g \in \mathcal{G},
\end{split}
\end{align}
where ReLU keeps prediction value $y_{g}^{'} \geq 0$. Then, the loss function can be defined as:
\begin{align}
\begin{split}
\mathit{Loss_p} = \frac{1}{\vert \mathcal{G} \vert} \sum_{g \in \mathcal{G}} \vert y_g - y_{g}^{'} \vert.\label{eq:lossp}
\end{split}
\end{align}

In addition, the MAE loss $\mathit{Loss}_l$ of LSTM of modeling time series in the competitive module can be calculated in the same way as Equation (\ref{eq:lossp}) could be jointly trained with $\mathit{Loss}_p$, since the more accurate states fitting of the published nodes leads to the more effective information aggregation.
Next, we adopt trade-off to assign the weights between $\mathit{Loss}_l$ and $\mathit{Loss}_p$, $\Theta$ represents the parameters of our model, so the objective function as follows:
\begin{align}
\begin{split}
\mathit{f}(\Theta) = \min_{\Theta} \left(\eta \mathit{Loss}_p + \left(1-\eta\right) \mathit{Loss}_l\right).
\end{split}
\end{align}

Finally, we use Stochastic Gradient Decent(SGD) to update parameters $\Theta$ of the model and apply exponential decay to the initial learning rate 0.02.

\section{Experiments}
In this section, we first introduce the dataset used in experiments. Then, the experimental setup is expressed in detail. Finally, we focus on evaluating the performances of GME in report of experimental results.
\subsection{Dataset Description}
Specifically, we conduct experiments on Indiegogo dataset \cite{liu2017enhancing} to verify the effectiveness of our model in the task of early fundraising estimating. There are $N = 14,143$ projects and $\vert S \vert = 1,862,097$ investing records in the dataset. And the dataset contains a variety of heterogeneous information about projects as shown in Table~\ref{table:features} mentioned in preliminaries. To test the generalization ability of our model on different scale dataset, we individually sample three sub datasets consisting of different amounts of projects (7K, 10K and 14K) from the complete dataset. In our experiments, we properly preprocess the dataset and divide it into training set and test set in chronological order with a split ratio of 5:1. To build historical market data for each target set, we use moving partition strategy to get inputs of model with a sliding time window above the time order. The sliding step is the segment periods mentioned above, and the length $t_h = \{1,2,...,7\}$ of time window would be adopted in experiments.
\subsection{Experimental Setup}
\subsubsection{Evaluation Protocols.}
In our study, we evaluated the prediction performance of our approach and the baselines through Root Mean Squared Error (RMSE) and Mean Absolute Error (MAE), which are the two standard evaluation metrics for prediction tasks \cite{wang2018learning}. Specifically, $\mathcal{G}$ represents a set includes target projects from one time window, for the project $i \in \mathcal{G}$ whose real target value $y_i$ and predicted value $\widetilde{y_i}$ described in problem overview, the evaluation protocols denoted as:
\begin{align}
\begin{split}
\mathit{RMSE} &= \sqrt{\frac{1}{\vert \mathcal{G} \vert} \sum_{i=0}^{\vert \mathcal{G} \vert}\left(y_i - \widetilde{y_i}\right)^{2}} ,\\
\mathit{MAE} &= \frac{1}{\vert \mathcal{G} \vert} \sum_{i=0}^{\vert \mathcal{G} \vert} {\vert y_i - \widetilde{y_i} \vert} .
\end{split}
\end{align}
\subsubsection{Implementation Details.}
For numerical features in static features, we convert them into the categorical type (one-hot encoder) by dispersion. Specifically, the dimensions of declared funding duration and declared pledged goal are 4 and 16. For text data, we first use word segmentation to deal with it. Then we remove all punctuations, convert all words to lowercase, and only keep words that appear more than 5 times. On this basis, we implement the doc2vec method to generate 50-dimensional vector for each description. For all models (include baselines) in our experiments, the size of the hidden state in LSTM and the state of nodes in graph are both 50, the dimension of FC layers are also 50, and the dropout after gated module are kept 0.9 empirically. The weight $\eta$ of trade-off is set as 0.7.

\subsubsection{Baselines.}
Since \textit{Early Fundraising Performance Estimating} is an open issue, many sequential methods in crowdfunding are not applicable to this task. Therefore, we compare the following well-designed models which are selected from the most related research fields, and we fine-tune some models that need to be transferred to adapt to our task.
\begin{itemize}
\item $\textbf{RFR}$ (\textit{Random Forest Regression}) is a regression-based method that can be used to model hand-crafted features \cite{liaw2002classification}. We process static features for prediction by the RFR whose max depth is 4.
\item $\textbf{MLP}$ (\textit{Multiple Layer Perceptron}) is a typical feedforward neural network trained with back propagation \cite{zhang1998comparison}. For target project $g$, we first concatenate the static feature $x_g$, the average pooling vector of running projects' competitiveness at $T_g$ and the average pooling vector of the initial state of nodes in $tree_g$ together. Then, these concatenated vectors are fed into an MLP with three FC layers for predictions.
\item $\textbf{LSTM}$ (\textit{Long Short-Term Memory networks}) is a recurrent neural network architecture that can avoid vanishing gradient, which is a useful tool for modeling sequence. We set its step size to 15 in compared experiments~\cite{hochreiter1997long}.
\item $\textbf{CLSTM}$ (\textit{Contextual LSTM}) takes the contextual information into account during the message propagation process \cite{ghosh2016contextual}, we create the contextual vector by discretized temporal variables in hours.
\item $\textbf{DTCN}$ (\textit{Deep Temporal Context Network}) is designed for sequential image popularity prediction \cite{wu2017sequential}. This model learns the long-term and short-term effect of context from historical data.
\item $\textbf{DMC}$ (\textit{Dynamic Market Competition}) can capture the competitiveness of projects in crowdfunding \cite{lin2018modeling}, we fine-tune the model slightly to make it applicable to our problem scenario.
\end{itemize}
In addition, we conduct the ablation experiments to verify the effectiveness of the main components of GME.
\begin{itemize}
\item $\textbf{GME-C}$ is refer to our complete model GME without the market evolution module. We compare it with other models to test the competition module.
\item $\textbf{GME-H}$ is refer to our complete model GME without the project competition module. We compare it with other models to test the market evolution module.
\end{itemize}
\linespread{1.26}
\begin{table}[t]
\large
\caption{Comparisons on the datasets.}
\centering
\resizebox{.95\columnwidth}{!}{
\begin{tabular}{c|cc|cc|cc}
\hline
\multirow{2}{*}{Models} & \multicolumn{2}{c|}{Indiegogo-7K} & \multicolumn{2}{c|}{Indiegogo-10K} & \multicolumn{2}{c}{Indiegogo-14K}\\

\cline{2-7} & MAE & RMSE & MAE & RMSE & MAE & RMSE\\
\hline
RFR & 0.2175 & 0.3343 & 0.2224 & 0.3293  & 0.2251 & 0.3376\\
MLP & 0.2201 & 0.3377 &0.2209 & 0.3351  & 0.2193 & 0.3305\\
LSTM & 0.2134 & 0.3316 &0.2238 & 0.3324 & 0.2117 & 0.3240\\
CLSTM & 0.2112 & 0.3279 &0.2128 & 0.3266 & 0.2146 & 0.3293\\
DTCN & 0.2024 & 0.3165 &0.1978 &0.3018  & 0.2044 & 0.3142\\
DMC & 0.1970 & 0.3100 &0.1956 & 0.3117 & 0.1933 & 0.2938\\
\hline
GME-C & 0.1941 & 0.3047 &0.1974 & 0.3141 & 0.2053 & 0.3078\\
GME-H & 0.1880 & 0.3086 &0.1930 & 0.3023 & 0.1914 & 0.2924\\
\textbf{GME} & \textbf{0.1771} & \textbf{0.2940} &\textbf{0.1819} & \textbf{0.2894} & \textbf{0.1836} & \textbf{0.2857}\\
\hline
\end{tabular}
}
\label{table:compar}
\end{table}

\subsection{Experimental Results}
\subsubsection{Model Comparison.}
The comparison results on the fundraising prediction task with different models including GME are shown in Table~\ref{table:compar}.
First, we focus on the performances of our complete approach GME, it achieves the best prediction performances on three data size settings with two evaluation metrics, and the minimum MAE is 0.1771, the minimum RMSE is 0.2857.
Compared with non-GME models (except GME-H and GME-C), GME outperforms other prediction algorithms with averages of 15.6\%, 14.0\% and 12.9\% relative MAE improvements on three data size settings.
Next, we can easily see that the components of GME are also able to get superior results. Especially, GME-C surpasses all models except GME on RMSE on Indiegogo-7K. And compared with three sequential models (LSTM, CLSTM, DTCN), the propagation tree structure of GME-H is not complicated but can effectively capture the evolution of the entire market environment. It leads to GEM-H gets more accurate results on MAE and RMSE.
Finally, observing the performance of the combination of modules, there is an obvious appearance that GME can achieve higher accuracy than GME-H and GME-C. It proves that combining competition learning with market evolution learning can further boost the performance.
To sum up, the effectiveness of market modeling in our approach is verified in the experiments.

\subsubsection{Influence of History Length.}
\begin{figure}[t]	
	\centering
	\begin{subfigure}[t]{0.46\columnwidth}
		\centering
		\includegraphics[width=\columnwidth]{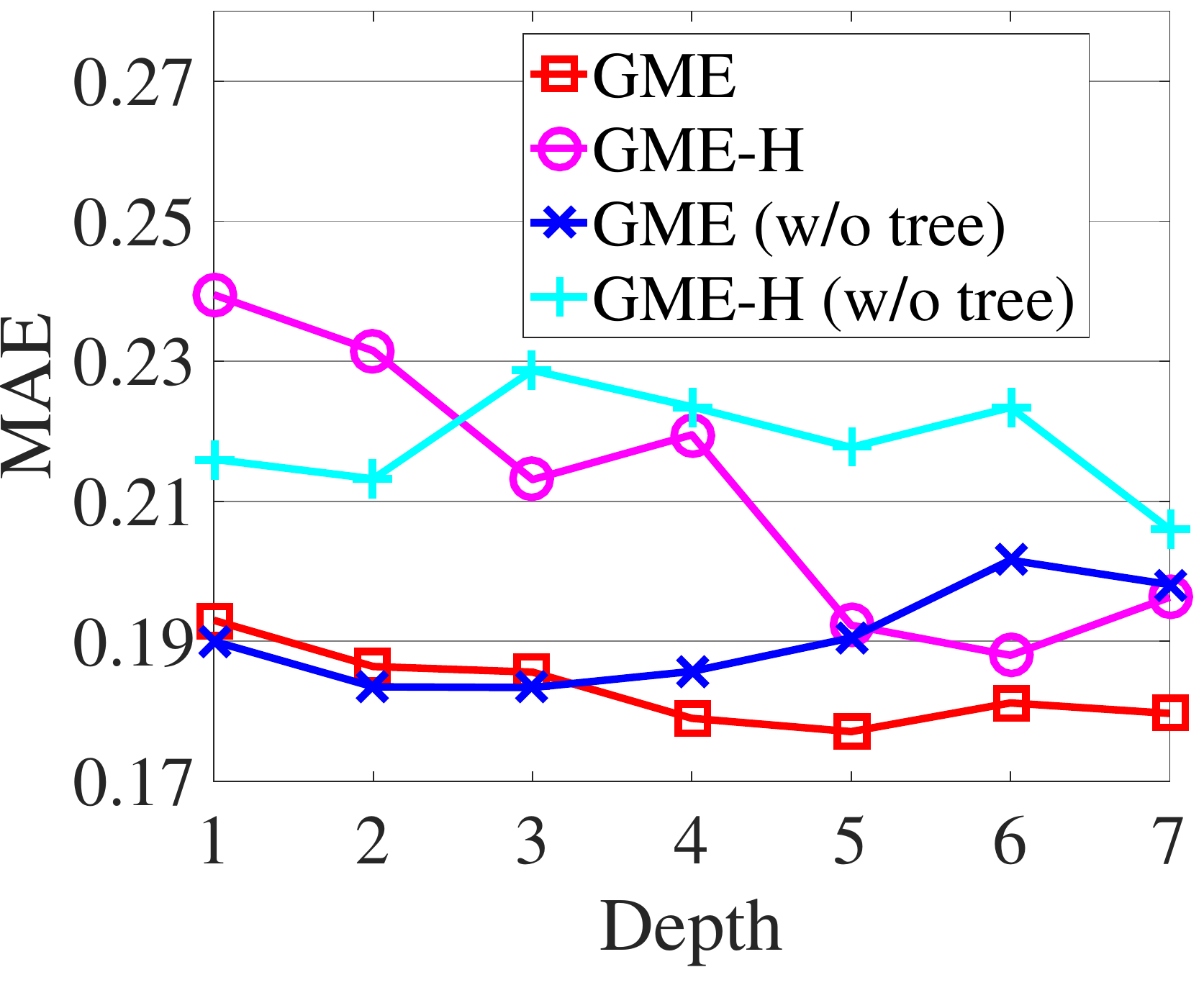}
		\caption{MAE changes with depth.}\label{fig:lensa}		
	\end{subfigure}
	\quad
	\begin{subfigure}[t]{0.46\columnwidth}
		\centering
		\includegraphics[width=\columnwidth]{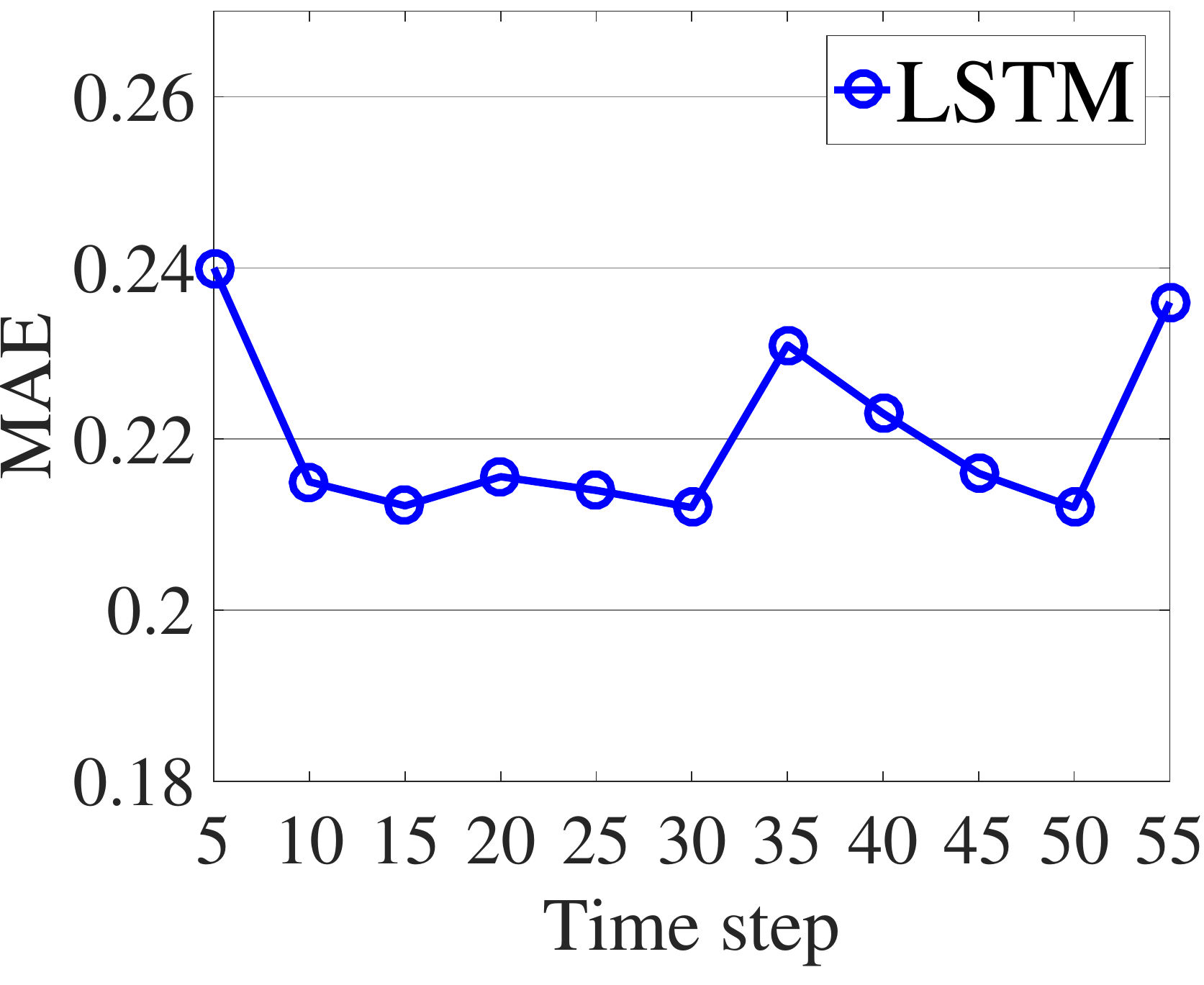}
		\caption{Timestep influence LSTM.}\label{fig:lensb}
	\end{subfigure}
	\caption{Influence of history length.}\label{fig:lens}
\end{figure}

In this section, first, we visualize the performance variations with the increase of the depth of propagation tree on Indiegogo-7K. The depth is the number of historical days we can observe before the pre-startup time of the target project. What stands out in Figure~\ref{fig:lensa} is that the performances of GME and GME-H become better with deeper depth in the overall trend. However, at the tail of these curves, GME and GME-H show slight upward trends. These tests reveal that lengthening history length without limit does not necessarily lead to better experimental results, since it probably to bring more noise. In our experiment, the most appropriate depth for GME is 5, it achieves the best result 0.1771 on MAE. Second, compare with the classic sequential model, as shown in Figure~\ref{fig:lensb}, we find that LSTM does not improve significantly with the increase of step size (the number of historical projects). Finally, to verify that the tree structure with time levels is more efficient than directly aggregating all historical nodes by GNN, we define ``w/o tree'' to denote not using tree structure. And we give the results of corresponding architectures GME (w/o tree) and GME-H (w/o tree). Obviously, GME (w/o tree) and GME-H (w/o tree) do not perform well, the greatest scores of them are 0.1834 and 0.2060. And they do not improve by the increase of the history length.
\begin{figure}[t]
\centering
\includegraphics[width=0.95\columnwidth]{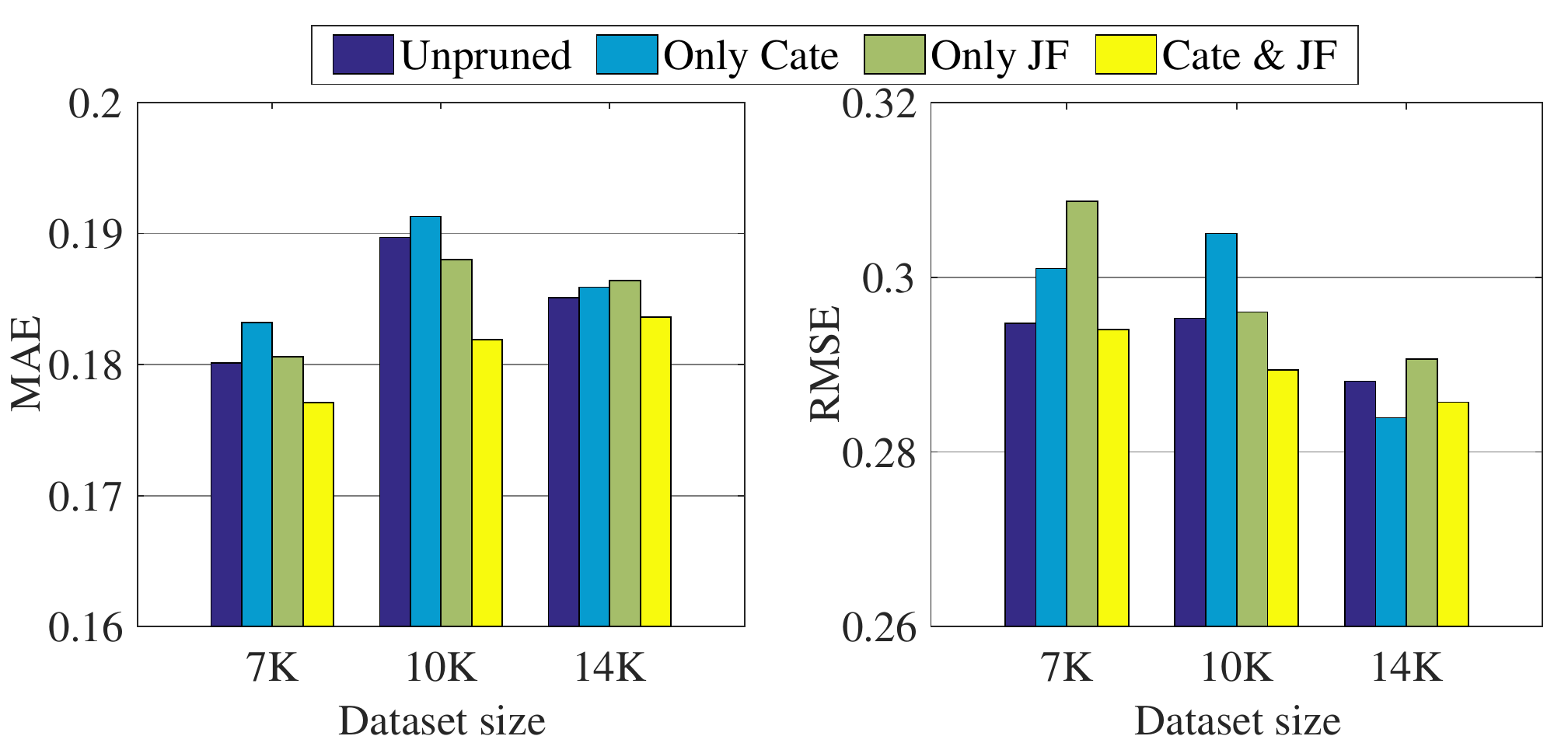}
\caption{Influence of pruning approaches.}
\label{fig:prun} 
\end{figure}

\subsubsection{Influence of Pruning Approaches.}
In the competitiveness module, we propose the pruning approach that retains only two physical relationships: projects belong to the same category and on the just-funded section.
The experimental results of pruning approaches are shown in Figure~\ref{fig:prun}, where all methods are based on GME. \textit{Unpruned} retains all edges, \textit{Only Cate} retains the edges between the target project and other nodes with the same category, \textit{Only JF} deletes all connections except the projects on the just-funded section at time $T_g$. Moreover, \textit{Cate \& JF} denotes the union of \textit{Only Cate} and \textit{Only JF}.
It can be seen in Figure~\ref{fig:prun}, pruning with only one physical relationship would reduce the accuracy of the model slightly, since it loses a considerable amount of information. But the significant result is that \textit{Cate \& JF} lead to a more accurate prediction than \textit{Unpruned} on our dataset. It proves that our pruning method is capable of avoiding more useless input for the influence aggregation.

\section{Conclusion}
In this paper, we studied the issue of estimating the early fundraising performance of the unpublished project on online crowdfunding platforms. Along this line, we proposed a Graph-based Market Environment model (GME) to explore the market environment and estimate the early performance of the target project. Specifically, we discriminatively modeled the market competition and market evolution by designing two graph-based neural network architectures and incorporating them into the joint optimization stage. We conducted extensive experiments on the real-world crowdfunding data collected from Indiegogo.com. The experimental results clearly validated the effectiveness of GME.

In the future, we will apply our models to other scenarios and applications leading to a more general framework.

\section*{Acknowledgements}
This research was partially supported by grants from the National Key Research and Development Program of China (No. 2016YFB1000904), and the National Natural Science Foundation of China (Grants No. U1605251, 61922073, 61672483). Qi Liu acknowledges the support of the Young Elite Scientist Sponsorship Program of CAST and the Youth Innovation Promotion Association of CAS (No. 2014299).

\bibliographystyle{aaai}
\bibliography{947-myreference}

\begin{thebibliography}{}

\bibitem[\protect\citeauthoryear{Bronstein \bgroup et al\mbox.\egroup
  }{2017}]{bronstein2017geometric}
Bronstein, M.~M.; Bruna, J.; LeCun, Y.; Szlam, A.; and Vandergheynst, P.
\newblock 2017.
\newblock Geometric deep learning: going beyond euclidean data.
\newblock {\em IEEE Signal Processing Magazine} 34(4):18--42.

\bibitem[\protect\citeauthoryear{Bruna \bgroup et al\mbox.\egroup
  }{2013}]{bruna2013spectral}
Bruna, J.; Zaremba, W.; Szlam, A.; and LeCun, Y.
\newblock 2013.
\newblock Spectral networks and locally connected networks on graphs.
\newblock {\em arXiv preprint arXiv:1312.6203}.

\bibitem[\protect\citeauthoryear{Cucurull, Taslakian, and
  Vazquez}{2019}]{cucurull2019context}
Cucurull, G.; Taslakian, P.; and Vazquez, D.
\newblock 2019.
\newblock Context-aware visual compatibility prediction.
\newblock In {\em Proceedings of the IEEE Conference on Computer Vision and
  Pattern Recognition},  12617--12626.

\bibitem[\protect\citeauthoryear{Defferrard, Bresson, and
  Vandergheynst}{2016}]{defferrard2016convolutional}
Defferrard, M.; Bresson, X.; and Vandergheynst, P.
\newblock 2016.
\newblock Convolutional neural networks on graphs with fast localized spectral
  filtering.
\newblock In {\em Advances in neural information processing systems},
  3844--3852.

\bibitem[\protect\citeauthoryear{Feng \bgroup et al\mbox.\egroup
  }{2019}]{feng2019temporal}
Feng, F.; He, X.; Wang, X.; Luo, C.; Liu, Y.; and Chua, T.-S.
\newblock 2019.
\newblock Temporal relational ranking for stock prediction.
\newblock {\em ACM Transactions on Information Systems (TOIS)} 37(2):27.

\bibitem[\protect\citeauthoryear{Ghosh \bgroup et al\mbox.\egroup
  }{2016}]{ghosh2016contextual}
Ghosh, S.; Vinyals, O.; Strope, B.; Roy, S.; Dean, T.; and Heck, L.
\newblock 2016.
\newblock Contextual lstm (clstm) models for large scale nlp tasks.
\newblock {\em arXiv preprint arXiv:1602.06291}.

\bibitem[\protect\citeauthoryear{Gilmer \bgroup et al\mbox.\egroup
  }{2017}]{gilmer2017neural}
Gilmer, J.; Schoenholz, S.~S.; Riley, P.~F.; Vinyals, O.; and Dahl, G.
\newblock 2017.
\newblock Neural message passing for quantum chemistry.
\newblock In {\em Proceedings of the 34th International Conference on Machine
  Learning-Volume 70},  1263--1272.
\newblock JMLR. org.

\bibitem[\protect\citeauthoryear{Gori, Monfardini, and
  Scarselli}{2005}]{gori2005new}
Gori, M.; Monfardini, G.; and Scarselli, F.
\newblock 2005.
\newblock A new model for learning in graph domains.
\newblock In {\em Proceedings. 2005 IEEE International Joint Conference on
  Neural Networks, 2005.}, volume~2,  729--734.
\newblock IEEE.

\bibitem[\protect\citeauthoryear{Hochreiter and
  Schmidhuber}{1997}]{hochreiter1997long}
Hochreiter, S., and Schmidhuber, J.
\newblock 1997.
\newblock Long short-term memory.
\newblock {\em Neural computation} 9(8):1735--1780.

\bibitem[\protect\citeauthoryear{Janku and
  Kucerova}{2018}]{janku2018successful}
Janku, J., and Kucerova, Z.
\newblock 2018.
\newblock Successful crowdfunding campaigns: The role of project specifics,
  competition and founders’ experience.
\newblock {\em Czech Journal of Economics and Finance (Finance a uver)}
  68(4):351--373.

\bibitem[\protect\citeauthoryear{Jin \bgroup et al\mbox.\egroup
  }{2019}]{jin2019estimating}
Jin, B.; Zhao, H.; Chen, E.; Liu, Q.; and Ge, Y.
\newblock 2019.
\newblock Estimating the days to success of campaigns in crowdfunding: A deep
  survival perspective.
\newblock In {\em Thirty-Third AAAI Conference on Artificial Intelligence}.

\bibitem[\protect\citeauthoryear{Kipf and Welling}{2016}]{kipf2016semi}
Kipf, T.~N., and Welling, M.
\newblock 2016.
\newblock Semi-supervised classification with graph convolutional networks.
\newblock {\em arXiv preprint arXiv:1609.02907}.

\bibitem[\protect\citeauthoryear{Le and Mikolov}{2014}]{le2014distributed}
Le, Q., and Mikolov, T.
\newblock 2014.
\newblock Distributed representations of sentences and documents.
\newblock In {\em International conference on machine learning},  1188--1196.

\bibitem[\protect\citeauthoryear{Li \bgroup et al\mbox.\egroup
  }{2015}]{li2015gated}
Li, Y.; Tarlow, D.; Brockschmidt, M.; and Zemel, R.
\newblock 2015.
\newblock Gated graph sequence neural networks.
\newblock {\em arXiv preprint arXiv:1511.05493}.

\bibitem[\protect\citeauthoryear{Li, Rakesh, and Reddy}{2016}]{li2016project}
Li, Y.; Rakesh, V.; and Reddy, C.~K.
\newblock 2016.
\newblock Project success prediction in crowdfunding environments.
\newblock In {\em Proceedings of the Ninth ACM International Conference on Web
  Search and Data Mining},  247--256.
\newblock ACM.

\bibitem[\protect\citeauthoryear{Liaw, Wiener, and
  others}{2002}]{liaw2002classification}
Liaw, A.; Wiener, M.; et~al.
\newblock 2002.
\newblock Classification and regression by randomforest.
\newblock {\em R news} 2(3):18--22.

\bibitem[\protect\citeauthoryear{Lin, Yin, and Lee}{2018}]{lin2018modeling}
Lin, Y.; Yin, P.; and Lee, W.-C.
\newblock 2018.
\newblock Modeling dynamic competition on crowdfunding markets.
\newblock In {\em Proceedings of the 2018 World Wide Web Conference},
  1815--1824.

\bibitem[\protect\citeauthoryear{Liu \bgroup et al\mbox.\egroup
  }{2017}]{liu2017enhancing}
Liu, Q.; Wang, G.; Zhao, H.; Liu, C.; Xu, T.; and Chen, E.
\newblock 2017.
\newblock Enhancing campaign design in crowdfunding: A product supply
  optimization perspective.
\newblock In {\em IJCAI},  695--702.

\bibitem[\protect\citeauthoryear{Tai, Socher, and
  Manning}{2015}]{tai2015improved}
Tai, K.~S.; Socher, R.; and Manning, C.~D.
\newblock 2015.
\newblock Improved semantic representations from tree-structured long
  short-term memory networks.
\newblock {\em arXiv preprint arXiv:1503.00075}.

\bibitem[\protect\citeauthoryear{Veli{\v{c}}kovi{\'c} \bgroup et
  al\mbox.\egroup }{2017}]{velivckovic2017graph}
Veli{\v{c}}kovi{\'c}, P.; Cucurull, G.; Casanova, A.; Romero, A.; Lio, P.; and
  Bengio, Y.
\newblock 2017.
\newblock Graph attention networks.
\newblock {\em arXiv preprint arXiv:1710.10903}.

\bibitem[\protect\citeauthoryear{Wang \bgroup et al\mbox.\egroup
  }{2018}]{wang2018learning}
Wang, W.; Zhang, W.; Wang, J.; Yan, J.; and Zha, H.
\newblock 2018.
\newblock Learning sequential correlation for user generated textual content
  popularity prediction.
\newblock In {\em IJCAI},  1625--1631.

\bibitem[\protect\citeauthoryear{Wu \bgroup et al\mbox.\egroup
  }{2017}]{wu2017sequential}
Wu, B.; Cheng, W.-H.; Zhang, Y.; Huang, Q.; Li, J.; and Mei, T.
\newblock 2017.
\newblock Sequential prediction of social media popularity with deep temporal
  context networks.
\newblock {\em arXiv preprint arXiv:1712.04443}.

\bibitem[\protect\citeauthoryear{Zhang \bgroup et al\mbox.\egroup
  }{1998}]{zhang1998comparison}
Zhang, Z.; Lyons, M.; Schuster, M.; and Akamatsu, S.
\newblock 1998.
\newblock Comparison between geometry-based and gabor-wavelets-based facial
  expression recognition using multi-layer perceptron.
\newblock In {\em Proceedings Third IEEE International Conference on Automatic
  face and gesture recognition},  454--459.
\newblock IEEE.

\bibitem[\protect\citeauthoryear{Zhang \bgroup et al\mbox.\egroup
  }{2019}]{zhang2019personalized}
Zhang, L.; Zhang, X.; Cheng, F.; Sun, X.; and Zhao, H.
\newblock 2019.
\newblock Personalized recommendation for crowdfunding platform: A
  multi-objective approach.
\newblock In {\em 2019 IEEE Congress on Evolutionary Computation (CEC)},
  3316--3324.
\newblock IEEE.

\bibitem[\protect\citeauthoryear{Zhao \bgroup et al\mbox.\egroup
  }{2017a}]{zhao2017sequential}
Zhao, H.; Liu, Q.; Zhu, H.; Ge, Y.; Chen, E.; Zhu, Y.; and Du, J.
\newblock 2017a.
\newblock A sequential approach to market state modeling and analysis in online
  p2p lending.
\newblock {\em IEEE Transactions on Systems, Man, and Cybernetics: Systems}
  48(1):21--33.

\bibitem[\protect\citeauthoryear{Zhao \bgroup et al\mbox.\egroup
  }{2017b}]{zhao2017tracking}
Zhao, H.; Zhang, H.; Ge, Y.; Liu, Q.; Chen, E.; Li, H.; and Wu, L.
\newblock 2017b.
\newblock Tracking the dynamics in crowdfunding.
\newblock In {\em Proceedings of the 23rd ACM SIGKDD International Conference
  on Knowledge Discovery and Data Mining},  625--634.
\newblock ACM.

\bibitem[\protect\citeauthoryear{Zhao \bgroup et al\mbox.\egroup
  }{2019}]{zhao2019voice}
Zhao, H.; Jin, B.; Liu, Q.; Ge, Y.; Chen, E.; Zhang, X.; and Xu, T.
\newblock 2019.
\newblock Voice of charity: Prospecting the donation recurrence \& donor
  retention in crowdfunding.
\newblock {\em IEEE Transactions on Knowledge and Data Engineering}.

\bibitem[\protect\citeauthoryear{Zhou, Huang, and
  Sch{\"o}lkopf}{2007}]{zhou2007learning}
Zhou, D.; Huang, J.; and Sch{\"o}lkopf, B.
\newblock 2007.
\newblock Learning with hypergraphs: Clustering, classification, and embedding.
\newblock In {\em Advances in neural information processing systems},
  1601--1608.

\end{thebibliography}

\end{document}